\pdfoutput=1

\documentclass[11pt]{article}

\newcommand{\draftonly}[1]{#1}
\renewcommand{\draftonly}[1]{}

\usepackage[final,hyperref]{acl}
\usepackage{times}
\usepackage{latexsym}

\usepackage[T1]{fontenc}
\usepackage[utf8]{inputenc}
\usepackage{microtype}

\usepackage{microtype}

\usepackage{amssymb,bm,mathtools,color}

\usepackage[normalem]{ulem}
\usepackage{enumitem}

\usepackage{booktabs}
\usepackage{multirow}
\usepackage{tabularx}
\usepackage{multicol}
\usepackage{multirow}
\usepackage{colortbl}
\usepackage{hhline}
\usepackage{stfloats}

\newcolumntype{Y}{>{\centering\arraybackslash}X}

\usepackage{xcolor}
\definecolor{red}{rgb}{0.74,0.08,0.10}
\definecolor{green}{rgb}{0.26,0.49,0.18}
\definecolor{blue}{rgb}{0.22,0.53,0.75}

\usepackage{colortbl}
\definecolor{Gray}{gray}{0.9}
\definecolor{LightCyan}{rgb}{0.75,1,1}

\makeatletter
\newcommand\notsotiny{\@setfontsize\notsotiny\@viiipt\@ixpt}
\makeatother

\usepackage{booktabs}
\usepackage{multirow}
\usepackage{graphicx}

\usepackage{pifont}
\usepackage{amsmath}
\usepackage{amssymb}
\usepackage{ifthen}
\usepackage{dsfont}

\usepackage{dcolumn}
\usepackage{siunitx}
\usepackage{makecell,rotating,array}

\usepackage{listings}
\usepackage{xcolor}
\usepackage{ltablex}
\usepackage{longtable}
\usepackage{multicol}
\usepackage{floatrow}

\usepackage{tikz}
\usetikzlibrary{bayesnet}
\usepackage{subfig}
\usepackage[linguistics]{forest}

\usepackage[ruled,vlined]{algorithm2e}

\usepackage{caption}

\usepackage{cancel}

\usepackage{xspace}
\usepackage{comment}
\usepackage{color,soul}

\usepackage{float}

\usepackage[noabbrev,capitalize]{cleveref} %

\crefname{page}{page}{pages}
\crefname{footnote}{footnote}{footnotes}   %
\crefname{equation}{equation}{equations}   %
\crefname{line}{line}{lines}               %
\crefrangeformat{line}{lines #3#1#4--#5#2#6}
\crefname{lstlsting}{Listing}{Listings}   
\crefname{section}{\S}{\S\S}
\Crefname{section}{\S}{\S\S}    %
\crefformat{section}{#2\S#1#3}  %
\Crefformat{section}{#2\S#1#3}
\crefrangeformat{section}{\S\S#3#1#4--#5#2#6}
\Crefrangeformat{section}{\S\S#3#1#4--#5#2#6}
\crefmultiformat{section}{\S#2#1#3}{ and~\S#2#1#3}{, \S#2#1#3}{ and~\S#2#1#3}
\Crefmultiformat{section}{\S#2#1#3}{ and~\S#2#1#3}{, \S#2#1#3}{ and~\S#2#1#3}
\crefrangemultiformat{section}{\S\S#3#1#4--#5#2#6}{ and~\S\S#3#1#4--#5#2#6}{, \S\S#3#1#4--#5#2#6}{ and~\S\S#3#1#4--#5#2#6}
\Crefrangemultiformat{section}{\S\S#3#1#4--#5#2#6}{ and~\S\S#3#1#4--#5#2#6}{, \S\S#3#1#4--#5#2#6}{ and~\S\S#3#1#4--#5#2#6}

\newcommand{\prob}[2][]{p\ifthenelse{\not\equal{}{#1}}{_{#1}}{}(#2)} %
\newcommand{\expect}[2][]{\text{\bf E}\ifthenelse{\not\equal{}{#1}}{_{#1}}{}\!\left[#2\right]}
\newcommand{\var}[2][]{\text{\bf Var}\ifthenelse{\not\equal{}{#1}}{_{#1}}{}\!\left[#2\right]}

\DeclareMathOperator{\argmax}{argmax}

\newcommand{\R}{\mathbb{R}}                     %

\newcommand{\eqdef}{\mathbin{\stackrel{\rm def}{=}}}

\newcommand{\defn}[1]{{\bf{#1}}}

\newcommand{\makename}[3][s]{%
  \expandafter\newcommand\csname #2\endcsname{#3\xspace}%
  \expandafter\newcommand\csname #2s\endcsname{#3#1\xspace}%
}

\makename{mice}{model-internal confidence estimators}
\makename{miceCaps}{Model-Internal Confidence Estimators}
\makename{miceTitle}{Model-Internal Confidence Estimation}

\makename{logitLens}{logit lens}

\makename{metricName}{expected tool-calling utility}
\makename{metricNameFirstCaps}{Expected tool-calling utility}
\makename{metricNameCaps}{Expected Tool-Calling Utility}
\makename{metricNameAbbr}{ETCU}
\newcommand{\AUC}{AUC--\metricNameAbbr}

\newcommand{\kernelReg}{NWKR}

\newcommand{\inputUtterance}[1]{\textit{#1}}
\newcommand{\outputToolCall}[1]{\texttt{#1}}

\newcommand{\tp}{\text{tp}}
\newcommand{\tn}{\text{tn}}
\newcommand{\fp}{\text{fp}}
\newcommand{\fn}{\text{fn}}

\newcommand{\numLayers}{\ensuremath{\ell}}

\newcommand{\hiddenStateAtLayer}[1]{H^{(#1)}}
\newcommand{\hiddenTokenAtLayer}[2]{{\bf h}^{(#1)}_{#2}}
\newcommand{\outputLayer}{\ensuremath{W_{\text{out}}}}

\title{%
    MICE for CATs: \\ 
    \miceTitle for Calibrating Agents with Tools
}

\author{\
Nishant Subramani$^\text{\ding{171}}$\thanks{\ \ Work performed during an internship at Microsoft.}%
\ \ \ \ \ 
\textbf{Jason Eisner$^\Diamond$}%
\ \ \ \ \ 
\textbf{Justin Svegliato$^\Diamond$}%
\ \ \ \ \
\\
\textbf{Benjamin Van Durme$^\Diamond$}%
\ \ \ \ \
\textbf{Yu Su$^{\Diamond\dagger}$}%
\ \ \ \ \ 
\textbf{Sam Thomson$^\Diamond$\thanks{\ \ Equal mentors.}}%
\\
\\$^\text{\ding{171}}$CMU LTI \hspace{1cm} $^\Diamond$Microsoft \\ 
$^\text{\ding{171}}$\texttt{nishant2@cs.cmu.edu} \\ %
$^\Diamond%
$\texttt{\{jason.eisner,jsvegliato,ben.vandurme,} \\ %
\texttt{yusu2,samuel.thomson%
\}@microsoft.com} \\
}

\begin{document}

\maketitle
\begin{abstract}
Tool-using agents that act in the world need to be both useful and safe.
Well-calibrated model confidences can be used to weigh the risk versus reward of potential actions,
but prior work shows that many models are poorly calibrated.
Inspired by interpretability literature exploring the internals of models, we propose a novel class of \textbf{\mice (MICE)} to better assess confidence when calling tools.
MICE first decodes from each intermediate layer of the language model using \textit{\logitLens}~\cite{logit_lens} and then computes similarity scores between each layer's generation and the final output.
These features are fed into a learned probabilistic classifier to assess confidence in the decoded output.
On the simulated trial and error (STE) tool-calling dataset using Llama3 models, we find that MICE beats or matches the baselines on smoothed expected calibration error.  Using MICE confidences to determine whether to call a tool significantly improves over strong baselines on a new metric, \textbf{expected tool-calling utility}. 
Further experiments show that MICE is sample-efficient, can generalize zero-shot to unseen APIs, and results in higher tool-calling utility in scenarios with varying risk levels.
Our code is open source, available at \url{https://github.com/microsoft/mice_for_cats}.

\end{abstract}

\section{Introduction}
\label{sec:intro}
Language models are increasingly being used
as tool-using agents, where they can generate executable API calls that can change external environments~\cite{schick2024toolformer,yan_berkeley-function-calling-leaderboard_2024,wang-etal-2024-llms-imaginarium,roy_benchclamp_2023}.
Sometimes the generated tool calls are relatively safe, and mistakes will have minimal impact %
(e.g., if \inputUtterance{"how many grand slams has Serena Williams won?"} resulted in the incorrect tool call
\outputToolCall{tennis\_reference\_count\_grand\_slams( name="venus williams")}, %
then the user would just be misinformed).
But other times, incorrect tool calls can be more harmful
(e.g., if \inputUtterance{"please remove slash.txt"} resulted in the incorrect tool call
\outputToolCall{cli(args="rm -rf /")},
then the user would lose the contents of their filesystem).

A \emph{confidence estimator} estimates the probability that another model's output is correct.
A simple confidence estimator for a language model would be based on the probability that the model itself assigns to its output (i.e., the product of token probabilities) or to its output's semantic equivalence class \cite{zhong-etal-2023-non,Farquhar2024}.  Yet prior work has shown that this method can be \emph{poorly calibrated}~\citep{jiang2021how,mielke-etal-2022-reducing, kadavath_language_2022, yin_large_2023}. %
A probabilistic classifier is \emph{well calibrated} if on an unseen test distribution, it is correct about as often as it thinks it is~\citep{Dawid1982TheWB,guo_calibration_2017,desai-durrett-2020-calibration, zhao_calibrate_2021,hashemi-et-al-2024}.  For example, of those unseen examples that it predicts to be positive with $\approx 25$\% probability, $\approx 25$\% really are positive. %
Well-calibrated probabilities can be used to guide downstream decisions, but calibration should never be one's only engineering target, as even a highly unsure classifier may be well-calibrated (see \cref{sec:metrics:ece}).

To that end, we introduce a class of \defn{\mice (MICE)} and an end-to-end metric, \defn{\metricName (\metricNameAbbr)}, to evaluate a tool-calling agent that consults a confidence estimator to decide when to launch the predicted tool call.\footnote{\label{fn:othertasks}We train and test confidence estimators specifically on the generation of \emph{tool calls}---a new setting for confidence estimation.
However, MICE could equally well be applied to well-studied confidence estimation settings in NLP, such as machine translation~\citep{blatz-etal-2004-confidence, Kumar2019CalibrationOE, wang-etal-2020-inference}, long-form generation~\citep{band2024linguistic}, and semantic parsing~\citep{ stengel-eskin-van-durme-2023-calibrated}.
}
MICE extracts features by decoding from the intermediate layers of a transformer-based large language model (LLM) and computes the similarities of those generations to the output of the final layer.
Based on these features and the LLM's raw confidence, it learns a model that outputs a confidence score.
MICE excels on {\metricNameAbbr}, increasingly outperforming two strong baselines as the cost of incorrect tool calls increases, without increasing calibration error.

This paper makes the following contributions:
    We propose a class of \mice (MICE) that are empirically well-calibrated on the task of assessing generated tool calls~(\S\ref{sec:mice}).
    We introduce a new metric, \metricName, that 
    combines accuracy and calibration to better evaluate %
    tool-calling agents~(\S\ref{sec:metrics}).
    Finally, we show that MICE is sample-efficient and can generalize to new tools, even in a zero-shot setting~(\S\ref{sec:results}). %

\section{\miceCaps} %
\label{sec:mice}

\begin{figure*}[t]
    \centering
    \includegraphics[width=\textwidth]{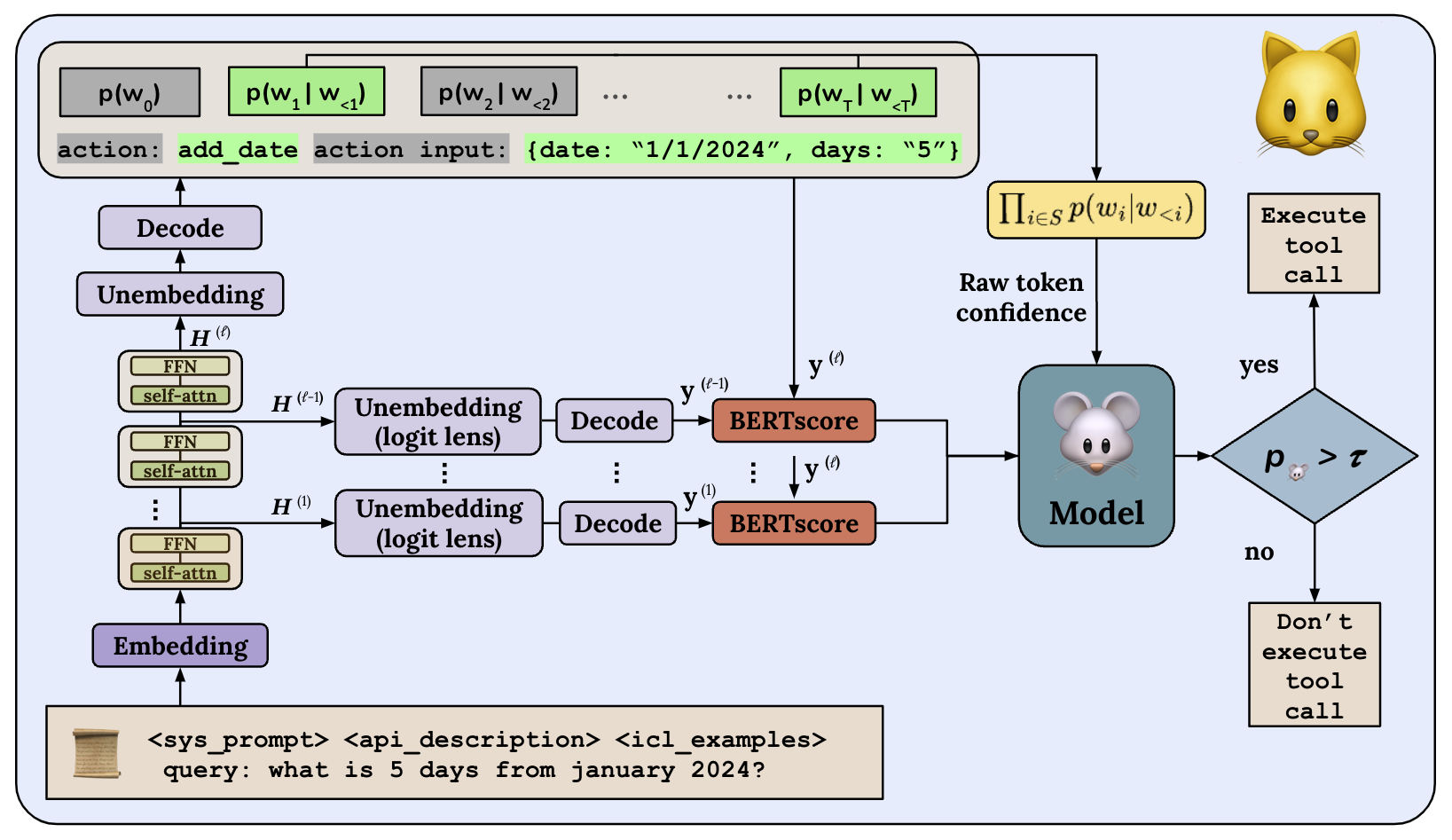}
    \caption{The MICE architecture.
    }
    \label{fig:mice-system-diagram}
\end{figure*}

MICE is a simple learned probabilistic classifier whose features are derived from model-internal activations.
Prior work on understanding the internals of
transformer language models has shown that intermediate layers at different depths encode different types of information, and that the activation spaces at various layers of these models can be nudged to generate sequences in targeted ways~\citep{tenney-etal-2019-bert, subramani2019can, subramani2020discovering, subramani-etal-2022-extracting, turner2023activation}.
Decoding from the layers of a transformer language model has provided insight into the underlying mechanisms and has been used in early-exit algorithms for faster generation~\citep{logit_lens, geva-etal-2022-transformer, schuster2022confident, Belrose2023ElicitingLP}.
For question answering tasks, %
decoding from roughly the first half of the layers of the language model produces unintelligible 
results,
but in later layers the model's predictions slowly refine into a plausible answer~\citep{merullo-etal-2024-language}.

We hypothesize that features from intermediate layers' hidden states could provide useful signal for calibration.
Drastic changes in the final few layers could indicate the inability for the LLM to pinpoint a tool call. 
As a result, we may trust a prediction that was slowly refined into an answer over the final 50\% of layers more than one that drastically changed in the final few layers, even if they had the same distribution at the end.

\begin{figure}[t]
    \centering
    \includegraphics[width=\linewidth]{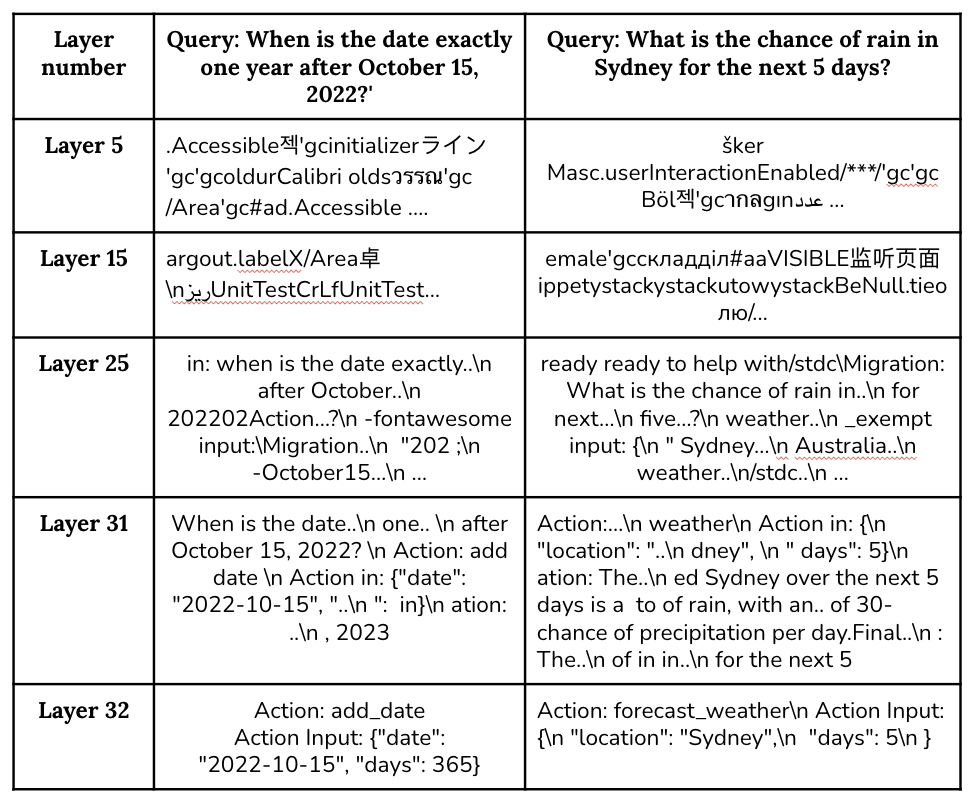}
    \caption{Example generations from the validation set across layers of the Llama3-8B-Instruct model.
    Generations from early layers (5, 15) are seemingly random, but later layers (25, 31) generate thematically relevant tokens.
    Layer 32 is the final layer.} %
    \label{fig:gens-per-layer}
\end{figure}

\paragraph{BERTScore Features} %
Since we hypothesize that intermediate layers' hidden states could be a useful signal for calibration beyond the confidences derived from the final layer, we decode from each layer, much as in \emph{logit lens}~\cite{logit_lens}, a model interpretation technique.
We first decode the output string ${\bf y}$ at temperature 0. This is the usual way to obtain model output in a task like tool calling.  Then at each layer $i < \ell$, we obtain a \emph{preliminary output string} ${\bf y}^{(i)}$ of the same length by per-token argmax decoding:\footnote{Note that taking $i=\ell$ in \labelcref{eqn:prelimy} would recover ${\bf y}$.  Because the transformer uses residual connections, each layerwise encoding $\hiddenTokenAtLayer{i}{t-1}$ has the same dimensionality $d$, so multiplication by the unembedding matrix \outputLayer is defined even when $i<\ell$.  All of these vector-matrix products can be computed in parallel by a matrix-matrix product, $\hiddenStateAtLayer{i}\,\outputLayer$ where $\hiddenStateAtLayer{i} \in \R^{\ell \times d}$.}
\begin{align}
    {\bf y}^{(i)}_t = \argmax \hiddenTokenAtLayer{i}{t-1}\,\outputLayer \label{eqn:prelimy}
\end{align}
where each 
$t$ is a token position in ${\bf y}$, and the row vector 
$\hiddenTokenAtLayer{i}{t-1} \in \R^{d}$ is the model's layer-$i$ encoding at the previous position, whose product with the unembedding matrix $\outputLayer \in \R^{d \times |V|}$ is a vector of logits $\in \R^{|V|}$.
Here, $d$ is the embedding size
and $|V|$ is the vocabulary size.
This results in $\numLayers$ strings%
, where $\numLayers$ is the number of layers of the model.

We then compute the BERTScore~\citep{zhang_bertscore_2019} between ${\bf y}$ and each ${\bf y}^{(i)}$.
These become the main input features to the MICE model.\footnote{\label{fn:altfeatures}BERTScore reencodes the strings ${\bf y}$ and ${\bf y}^{(i)}$ with a separate model (see \cref{sec:architectures}) and aligns their tokens.  We found that the alignments were not always trivial.  BERTscore performed significantly better than methods we explored initially, which compared the  $\mathrm{softmax}(\hiddenTokenAtLayer{i}{t-1}\,\outputLayer)$ distributions rather than argmax-decoding single strings ${\bf y}^{(i)}$.  See \cref{sec:limitations} for other options.}

\paragraph{Raw Confidence Feature} We also integrate the raw confidence of the language model in generating the tool call as a feature to the MICE model.
We calculate this by computing the product of the probabilities of the tokens in the generated tool call.
We notice that including formatting tokens, which are always present in the tool call, leads to increased noise and a less accurate estimate of confidence, so we 
omit the tokens associated with formatting. The gray tokens in Figure~\ref{fig:mice-system-diagram} were omitted, while the green ones were included.

\paragraph{Model Architecture}
We train a simple supervised classifier that predicts whether the generated tool call ${\bf y}$ is correct.
It maps from the input features—the BERTScores and the raw confidence—to a probability of correctness (i.e., a confidence estimate).
Any trainable model of this form could be used here; the specific architectures and baselines that we tried will be described in \cref{sec:architectures}.

\label{sec:methods}

\section{Metrics}
\label{sec:metrics}

Perhaps the most widely reported calibration metric is \defn{expected calibration error} (\S\ref{sec:metrics:ece}).  As mentioned in the introduction, however, minimizing ECE should not be our only goal.
We also introduce a utility metric, \defn{\metricName}~(\S\ref{sec:metrics:reward}), to assess the performance of a simple agent that makes call/no-call decisions by using our well-calibrated confidence estimates.
This metric 
is parameterized by the cost of false positives relative to the reward of true positives.

\subsection{Expected Calibration Error (ECE)}
\label{sec:metrics:ece}

\begin{figure}[t!]
    \centering
    \includegraphics[width=\linewidth]{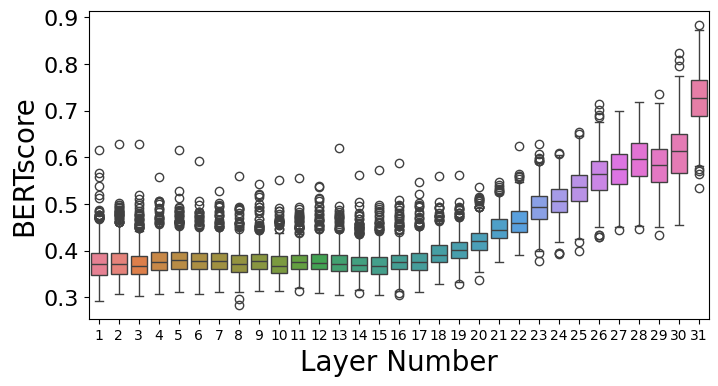}
    \caption{BERTScore similarities between the generated string $\bf y$ and the preliminary strings ${\bf y}^{(i)}$ from earlier layers, for Llama3-8B-Instruct on the STE validation set~\cite{wang-etal-2024-llms-imaginarium}. See also \cref{fig:bertscores-posneg} in \cref{sec:experimental-details}.   %
}
    \label{fig:bertscores-per-layer}
\end{figure}

Expected calibration error~\cite[\defn{ECE};][]{Naeini2014BinaryCC, Naeini2015ObtainingWC} is computed by constructing a histogram binned by predicted confidence, $\hat{p}$.
The accuracy of examples within a given bin is compared to the mean predicted confidence within that bin, $|\text{acc} - \bar{\hat{p}}|$.
These absolute differences are then averaged across bins, with each bin weighted by the fraction of examples in that bin.

We use a recently improved variant of ECE, \defn{smooth ECE}~\cite[\defn{smECE};][]{blasiok2024smooth}, which replaces histogram binning with  Nadaraya-Watson kernel regression~\citep{nadaraya1964estimating, watson1964smooth}. %
A reflected Gaussian kernel is used; the kernel width is determined automatically from the data, yielding a consistent estimator.

However, ECE and smECE 
do not distinguish between an \emph{oracle} classifier that returns $\hat{p}=1.0$ on correct outputs and $\hat{p}=0.0$ on incorrect outputs, and a \emph{maximally uninformative} probabilistic classifier that always predicts the base accuracy rate.  That is, if 70\% of all predictions are correct, then a trivial system that gave $\hat{p}=0.7$ on every example would be perfectly calibrated ($\text{ECE}=0$), yet mostly useless!
We would prefer a system that tends to return high $\hat{p}$ on correct tool calls and low $\hat{p}$ on incorrect ones, so that we can execute the former and avoid executing the latter.\looseness=-1

\label{sec:metrics:smoothece}

\vspace{-2pt}
\subsection{\metricNameCaps (\metricNameAbbr)}
\label{sec:metrics:reward}

We thus introduce a parameterized metric, \emph{\metricName}, which approximates actual utility in situations where a \emph{calibrated} confidence score $\hat{p}$ is used to decide of whether or not to execute a specific tool call generated by the language model.
We assume we know the expected utility for each of the four possible outcomes:
    \defn{$\tp > 0$} (true positive), for when the agent executes a \emph{correctly} predicted call;
    \defn{$\fp < 0$} (false positive), for when the agent executes an \emph{incorrectly} predicted call;
    \defn{$\tn \approx 0$} (true negative), for when the agent avoids executing an \emph{incorrect} call;
    \defn{$\fn \approx 0$} (false negative), for when the agent fails to execute a \emph{correct} call. %
    $\tn$ and $\fn$ may be slightly negative to account for time wasted making the unused prediction.  $\fp$ may be highly negative, e.g., if the agent erroneously deletes all of the user's documents, makes a large unintended purchase, or sends an offensive email.

The exact values of $\tp, \fp, \tn,$ and $\fn$ will depend on the specific task that the agent must perform, and could be assigned by a domain expert or learned from data, like human preferences~\cite{christiano2017deep}.\footnote{%
They will also often depend on the predicted API and arguments.
A more careful MBR practitioner would ideally condition on these and assign utilities to each possible pair (gold specific action, chosen specific action).  Our coarser expectations $\{\tp,\fp,\tn,\fn\}$ result in cruder decisions.}
Once these values have been assigned, the minimum Bayes risk~\citep[\textbf{MBR};][p.~27]{bickel_mathematical_1977} decision can be calculated;
it is to execute the predicted tool call if and only if the estimated confidence $\hat{p}$ is above the threshold

\noindent
\begin{align}
\hat{p} > \tau & \eqdef \frac{\tn - \fp}{(\tp - \fn) + (\tn - \fp)}
\label{eqn:tau}
\end{align}
Calibration ensures that of all predicted tool calls with confidence $\approx \hat{p}$, about $\hat{p}$ are correct.  
The decision rule \labelcref{eqn:tau} makes either \emph{all} such calls or \emph{none} of them, according to whether the expected utility per call is higher with \emph{all} ($\hat{p}\,\tp + (1-\hat{p})\,\fp$) or with \emph{none} ($\hat{p}\,\fn + (1-\hat{p})\,\tn$).  
The threshold $\tau$ is high ($> 0.5$) if avoiding bad calls (benefit $\tn-\fp$) is more important than executing good calls (benefit $\tp-\fn$).\looseness=-1

\paragraph{Normalizing:}
These four values can be scaled by any positive constant, %
and translated by any real constant, %
without affecting the optimal threshold or the utility (modulo that affine transform)~\cite{gleave2021quantifying}. %
That is, we can choose a measurement scale for our utilities (without loss of generality) such that $\tp = 1$ and $\fn = 0$.
Two degrees of freedom still remain ($\tn$ and $\fp$).
In most tool-using scenarios, $\tn$ will be extremely close to $\fn$, because in both cases the immediate action by the agent is the same (do not execute) and thus has the same effect regardless of the predicted action.\footnote{$\tn$ might differ from $\fn$ because \emph{subsequent} actions may diverge, and each utility should ideally include the expected future reward over all possible rollouts.
For example, it might be slightly easier to ask clarifying questions when the original prediction was correct (implying $\fn > \tn$ and raising $\tau$).}
If we further assume (\emph{with} loss of generality) that $\tn=\fn=0$,
the intuitive interpretation is that the agent gets 1 ``credit'' %
(an arbitrary utility unit) for completing its task, 0 credits for doing nothing (regardless of whether that was the best decision), and $\fp < 0$ credits for doing something wrong.
The single remaining degree of freedom $\fp$ \emph{is} the risk/utility ratio, defining how costly it is %
to attempt and fail.
In this (slightly less general) case, the MBR decision rule \labelcref{eqn:tau} simplifies to:
\begin{align}
    \hat{p} > \tau \eqdef \frac{-\fp}{1 + -\fp}
    = \frac{\fp}{\fp - 1}
\end{align}

\paragraph{Settings for \metricName:}
To understand how confidence estimators perform at different risk levels, we 
choose three different values of $\fp$ under which to measure normalized risk (\cref{tab:results}).
Each setting of $\fp$ determines a threshold $\tau$ that the Bayes-optimal policy will use.

\textit{High Risk:} Tasks where executing an incorrect tool call is much worse than the reverse error.
We choose $\fp=-9$ for this setting, giving $\tau = 0.9$.

\textit{Medium Risk:} For these tasks, executing an incorrect tool call is as bad as executing the correct tool call is good ($\fp = -\tp = -1$), giving $\tau = 0.5$.

\textit{Low Risk:} These are tasks where executing an incorrect tool call
has relatively low potential downside.
We choose $\fp=-\frac{1}{9}$, giving $\tau = 0.1$.

\begin{figure*}[t!]
\centering
\includegraphics[width=\linewidth]{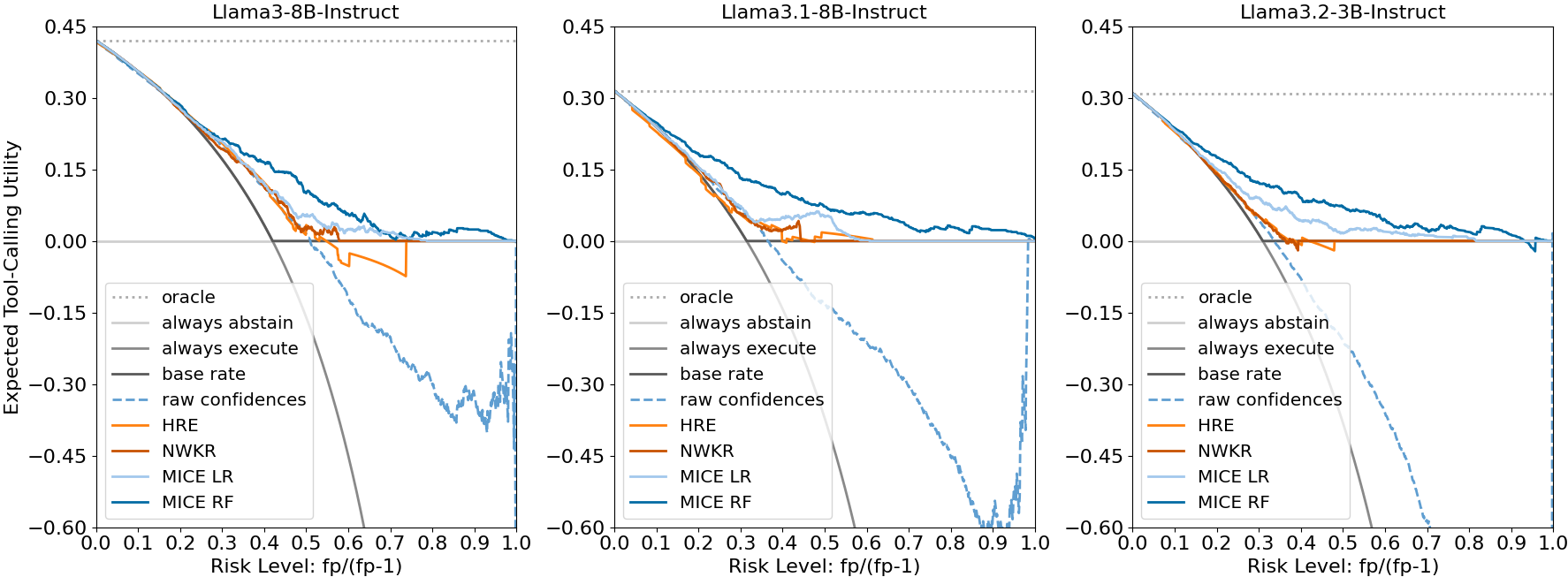}
\caption{\metricNameFirstCaps on the test set at varying risk levels.
We include four trivial policies for reference: \texttt{oracle} executes only when the underlying model is correct (an upper bound); \texttt{always abstain} never executes, getting reward 0; \texttt{always execute} never abstains; and the \texttt{base rate} policy switches from \texttt{always execute} to \texttt{always abstain} when the risk level exceeds the base accuracy.
All policies perform similarly at low risk levels, where \texttt{always execute} is close to optimal and hard to improve on.
MICE models show clear improvements in the medium and high risk regimes.
}
\label{fig:enr-plot-with-zeroshot}
\end{figure*}

\paragraph{Area Under Curve (AUC):}
More generally, we can compute an expected value for \emph{any} $\tau \in (0,1)$.  This yields an ``\metricName'' curve (Figure~\ref{fig:enr-plot-with-zeroshot}) for a given confidence estimator on a given dataset.
Any given applied setting may only be interested in a single $\tau$ along the curve.
Still, to compare estimators overall, it may be useful to consolidate the curve into a single number, summarizing an estimator's performance across all risk levels.
Taking inspiration from the area under the receiver operating characteristic (ROC) curve~\citep{marcum1960statistical}, we take the average of the \metricName values at every point along the curve, which can be regarded as the (signed) \defn{area under the curve} (\defn{AUC}).%
\footnote{%
In practice, we approximate AUC by evaluating \metricName at each $\tau \in \{0.001, 0.002, \ldots, 0.999\}$. %
}
Since our formulation sets $\tn=\fn=0$, always abstaining gets a \metricName score of $0$ regardless of risk level, and thus an AUC of $0$.
Because utilities can be negative, AUC values can also be negative.
This occurs when model is overconfident (due to poor calibration) in too many high-risk predictions.

\begin{table*}[t!]
\small
\newcolumntype{d}{r@{.}l}
\newcommand\mc[1]{\multicolumn{1}{c}{#1}}
\begin{tabular}{@{}l l c c c c c@{}}
\toprule
 & & & \multicolumn{4}{c}{\textbf{Tool-Calling Utility} ($\uparrow$)}\\
\cmidrule(r){4-7}
\textbf{Model} & \textbf{Confidence Estimator}                    & {\textbf{smECE} ($\downarrow$)} & {Low risk} & {Medium risk} & {High risk} & {AUC} \\
\midrule
\multirow{6}{*}%
 & {Raw Confidence}                      & 0.184 & \underline{0.351} & \textcolor{white}{*-}0.015\color{white}$^{*\dagger}$ & \textcolor{white}{*}-0.323\color{white}$^{*\dagger}$ & \textcolor{white}{*-}0.001\color{white}$^{*\dagger}$ \\
 & HRE & 0.041 & {\bf 0.356} & \textcolor{white}{*-}0.033\color{white}$^{*\dagger}$ & \textcolor{white}{*-}\underline{0.000}\color{white}$^{*\dagger}$ & \textcolor{white}{*-}0.110\color{white}$^{*\dagger}$  \\
 & \kernelReg & 0.039 & {\bf 0.356} & \textcolor{white}{*-}0.027\color{white}$^{*\dagger}$ & \textcolor{white}{*-}\underline{0.000}\color{white}$^{*\dagger}$ & \textcolor{white}{*-}0.118\color{white}$^{*\dagger}$  \\
\rowcolor{LightCyan}\cellcolor{white}
Llama3-8B-Instruct 
 & MICE LR Zeroshot  & {\bf 0.036} & {\bf 0.356} & \textcolor{LightCyan}{*-}0.025\color{LightCyan}$^{*\dagger}$ & \textcolor{LightCyan}{*-}\underline{0.000}\color{LightCyan}$^{*\dagger}$ & \textcolor{LightCyan}{*-}0.114\color{LightCyan}$^{*\dagger}$ \\
\rowcolor{LightCyan}\cellcolor{white}
 & MICE RF Zeroshot        & 0.065 & 0.349 & \textcolor{LightCyan}{*-}0.011\color{LightCyan}$^{*\dagger}$ & \textcolor{LightCyan}{*}-0.016\color{LightCyan}$^{*\dagger}$ & \textcolor{LightCyan}{*-}0.096\color{LightCyan}$^{*\dagger}$ \\
\rowcolor{LightCyan}\cellcolor{white}
 & MICE LR & \underline{0.037} & {\bf 0.356} & \textcolor{LightCyan}{*-}\underline{0.055}\color{LightCyan}$^{*\dagger}$ & \textcolor{LightCyan}{*-}\underline{0.000}\color{LightCyan}$^{*\dagger}$ & \textcolor{LightCyan}{*-}\underline{0.125}$^{*\color{LightCyan}\dagger}$ \\
\rowcolor{LightCyan}\cellcolor{white}
& {MICE RF} & 0.040 & {\bf 0.356} & \textcolor{LightCyan}{*-}{\bf 0.100}$^{\dagger\color{LightCyan}*}$ & \textcolor{LightCyan}{*-}{\bf 0.024}$^{*\dagger}$ & \textcolor{LightCyan}{*-}{\bf 0.144}$^{*\dagger}$ \\
\midrule
\multirow{7}{*}%
 & Raw Confidence                      & 0.229 & \underline{0.240} & \textcolor{white}{*}-0.129\color{white}$^{*\dagger}$ & \textcolor{white}{*}-0.579\color{white}$^{*\dagger}$ & \textcolor{white}{*}-0.127\color{white}$^{*\dagger}$ \\
 & HRE & \underline{0.050} & 0.229 & \textcolor{white}{*-}0.017\color{white}$^{*\dagger}$  & \textcolor{white}{*-}\underline{0.000}\color{white}$^{*\dagger}$    & \textcolor{white}{*-}0.060\color{white}$^{*\dagger}$  \\
 & \kernelReg & {\bf 0.032} & 0.239 & \textcolor{white}{*-}0.000\color{white}$^{*\dagger}$ & \textcolor{white}{*-}\underline{0.000}\color{white}$^{*\dagger}$ & \textcolor{white}{*-}0.063\color{white}$^{*\dagger}$ \\
\rowcolor{LightCyan}\cellcolor{white}
Llama3.1-8B-Instruct 
 & MICE LR Zeroshot  & 0.054 & 0.239 & \textcolor{LightCyan}{*-}0.019\color{LightCyan}$^{*\dagger}$ & \textcolor{LightCyan}{*-}\underline{0.000}\color{LightCyan}$^{*\dagger}$ & \textcolor{LightCyan}{*-}0.061\color{LightCyan}$^{*\dagger}$ \\
\rowcolor{LightCyan}\cellcolor{white}
 & MICE RF Zeroshot        & \underline{0.050} & 0.233 & \textcolor{LightCyan}{*-}\underline{0.051}$^{\dagger\color{LightCyan}*}$  & \textcolor{LightCyan}{*}-0.008\color{LightCyan}$^{*\dagger}$ & \textcolor{LightCyan}{*-}\underline{0.073}\color{LightCyan}$^{*\dagger}$ \\
\rowcolor{LightCyan}\cellcolor{white}
 & MICE LR & \underline{0.050} & 0.239 & \textcolor{LightCyan}{*-}\underline{0.051}$^{*\dagger}$  & \textcolor{LightCyan}{*-}\underline{0.000}\color{LightCyan}$^{*\dagger}$  & \textcolor{LightCyan}{*-}0.071$^{*\dagger}$ \\
\rowcolor{LightCyan}\cellcolor{white}
& MICE RF & {\bf 0.032} & {\bf 0.248} & \textcolor{LightCyan}{*-}{\bf 0.072}$^{*\dagger}$  & \textcolor{LightCyan}{*-}{\bf 0.023}$^{*\dagger}$  & \textcolor{LightCyan}{*-}{\bf 0.104}$^{*\dagger}$  \\
\midrule
\multirow{7}{*}%
 & Raw Confidence                      & 0.312 & 0.231 & \textcolor{white}{*}-0.209\color{white}$^{*\dagger}$ & \textcolor{white}{*}-1.528\color{white}$^{*\dagger}$ & \textcolor{white}{*}-0.458\color{white}$^{*\dagger}$ \\
 & HRE & 0.036 & 0.229 & \textcolor{white}{*-}0.000\color{white}$^{*\dagger}$    & \textcolor{white}{*-}\underline{0.000}\color{white}$^{*\dagger}$    & \textcolor{white}{*-}0.057\color{white}$^{*\dagger}$  \\
 & \kernelReg & \underline{0.027} & \underline{0.233} & \textcolor{white}{*-}0.000\color{white}$^{*\dagger}$ & \textcolor{white}{*-}\underline{0.000}\color{white}$^{*\dagger}$ & \textcolor{white}{*-}0.057\color{white}$^{*\dagger}$  \\
 \rowcolor{LightCyan}\cellcolor{white}
Llama3.2-3B-Instruct 
 & MICE LR Zeroshot  & 0.032 & 0.230 & \textcolor{LightCyan}{*-}0.008\color{LightCyan}$^{*\dagger}$ & \textcolor{LightCyan}{*-}\underline{0.000}\color{LightCyan}$^{*\dagger}$ & \textcolor{LightCyan}{*-}0.064\color{LightCyan}$^{*\dagger}$  \\
\rowcolor{LightCyan}\cellcolor{white}
 & MICE RF Zeroshot  & 0.061 & 0.232 & \textcolor{LightCyan}{*-}\underline{0.021}\color{LightCyan}$^{*\dagger}$ & \textcolor{LightCyan}{*}-0.068\color{LightCyan}$^{*\dagger}$ & \textcolor{LightCyan}{*-}0.049\color{LightCyan}$^{*\dagger}$ \\
\rowcolor{LightCyan}\cellcolor{white}
 & MICE LR & {\bf 0.025} & 0.232 & \textcolor{LightCyan}{*-}0.016\color{LightCyan}$^{*\dagger}$   & \textcolor{LightCyan}{*-}\underline{0.000}\color{LightCyan}$^{*\dagger}$    & \textcolor{LightCyan}{*-}\underline{0.073}$^{*\dagger}$  \\
\rowcolor{LightCyan}\cellcolor{white}
& MICE RF & 0.041 & {\bf 0.237} & \textcolor{LightCyan}{*-}{\bf 0.071}$^{*\dagger}$  & \textcolor{LightCyan}{*-}{\bf 0.013}\color{LightCyan}$^{*\dagger}$ & \textcolor{LightCyan}{*-}{\bf 0.095}$^{*\dagger}$ \\
\bottomrule
\end{tabular}
\caption{%
Results on the %
test set. %
Lower smECE is better, while higher tool-calling utility is better.
\textbf{Bold} indicates the best result in each category and \underline{underline} indicates the second best result in each category.
$^*$ indicates statistical significance compared to HRE and $\dagger$ indicates significance compared to \kernelReg\
(p-value < 0.05, permutation test).
}
\label{tab:results}
\end{table*}

\section{Experiments}
\label{sec:experiments}

We now look at training MICE and using it at test time to measure both smooth expected calibration error (smECE) and \metricName.

\subsection{Dataset}
Our experiments use the simulated trial-and-error (STE) dataset \cite{wang-etal-2024-llms-imaginarium}.
The dataset was synthetically generated by simulating plausible tool-using scenarios for a given API and using GPT3.5-turbo with execution feedback to identify (presumptively) correct tool calls.

The dataset consists of English-language queries that require calling 50 distinct APIs.  For tool call generation, we few-shot prompt an off-the-shelf LLM with examples from a \emph{demonstration set} consisting of 4,520 examples taken from the STE training set. An alternative would have been to fine-tune the LLM on this demonstration set.  

To train MICE, we use the rest of the STE training set, split into a \emph{training set} of 1500 examples (30 from each API) and a \emph{validation set} of 750 examples (15 from each API).  We then evaluate MICE on STE's \emph{test set} of 750 examples.  In all cases, we label a generated tool call as correct if and only if it exactly matches the one given by STE.\looseness=-1

\subsection{LLMs}

We consider %
three LLMs: Llama3-8B-Instruct, Llama3.1-8B-Instruct, and Llama3.2-3B-Instruct \citep{dubey2024llama}.
We build and evaluate a separate MICE classifier for each LLM.

\subsection{Experimental Settings}
We run each LLM on our validation and test sets in an 8-shot in-context learning setting, following~\citet{wang-etal-2024-llms-imaginarium}, using greedy decoding to generate the tool calls $\bf y$.
For each evaluation example, the 8 in-context learning examples are selected from our demonstration set according to the procedure of~\citet{wang-etal-2024-llms-imaginarium}, which computes similarity to the evaluation example using SentenceBERT~\citep{reimers-gurevych-2019-sentence}.

We train a baseline or MICE regressor on the training set to predict whether tool calls are correct, and use the validation set for hyperparameter and model selection. Features used by MICE regressors were described in \cref{sec:mice}.
We then evaluate the regressor on the test set, using the metrics of \cref{sec:metrics}.\looseness=-1

\subsection{MICE Configurations \& Baselines}\label{sec:architectures}
\paragraph{Raw Confidence} 
Our first baseline is the raw confidence score from \cref{sec:mice}, which can be used directly as a confidence estimate $\hat{p}$.  Recall that we defined this as $\prod_{i \in S} p(w_i | w_{<i})$, where $S$ is the subset of token indices that are relevant to the tool call.
$S$ omits the tokens associated with formatting (\outputToolCall{"action:"} and \outputToolCall{"action input:"}, which are generated for every tool call), and also omits tokens that are generated \emph{after} the arguments of the tool call.
We observed in initial experiments that including these irrelevant tokens resulted in worse calibration.
We also observed that taking the minimum probability across generated tokens instead of the joint probability (as in \citet{zhou-etal-2022-online, stengel-eskin-van-durme-2023-calibrated}) resulted in little effective difference.
Note that calculating raw confidence does not require any learning, so neither the training nor validation set is used.
Raw confidence is also used as a base feature in the estimators described below.

\paragraph{%
Histogram Regression Estimator \citep[HRE;][]{nobel_histogram_regression_1996}}
For our second (stronger) baseline, we use a standard method to calibrate the previous baseline.  We use the training set to
construct a histogram binned by raw confidence scores.
We use 25 bins: $[0,0.04), [0.04,0.08), \ldots, [0.96, 1.0]$.
To map from a raw confidence score $c$ to a recalibrated estimate $\hat{p}$, we look up $c$'s bin, and return the percentage of examples in that bin that are correct.
Note that this is the same histogram construction used to calculate traditional ECE (except here constructed on the training set), and so should be expected to perform well on ECE metrics.

\paragraph{Kernel Regressor (\kernelReg)}
Here, rather than using a histogram with fixed bins to 
recalibrate%
, we use Nadaraya–Watson kernel regression~\citep{nadaraya1964estimating, watson1964smooth}, following the exact procedure~\citet{blasiok2024smooth} used to compute smECE.
Analogously to above, since this follows the exact same procedure as in smECE, we should expect it to perform well under that metric.\footnote{HRE and \kernelReg\ learn to map a single confidence input feature to a recalibrated output confidence.
Any confidence estimator can be calibrated in this way on held-out data.
Other common approaches to this problem include isotonic regression and Platt scaling~(\S\ref{sec:related-work}).}

\paragraph{MICE Models}
We extract features as described in \cref{sec:mice}, using DeBERTa-xlarge-mnli to compute the BERTScore features as it is the strongest BERTScore base model~\citep{he2021deberta}. This gives $\ell-1$ BERTScore features along with the raw confidence feature.
There are $\ell= 32$ layers for Llama3 and 3.1 and 28 layers for Llama3.2.  

\textit{MICE Logistic Regressor (MICE LR):} We train a logistic regression model with an L2 regularization strength of 2 %
to predict whether the tool call is correct or not.

\textit{MICE Random Forest (MICE RF):}
We train a random forest classifier using 1000 trees each with a maximum depth of 20 and a maximum of 10 features to use at each split, using the Scikit-Learn package~\citep{scikit-learn}.
Other hyperparameters are set to defaults.
This model is also trained to predict whether the tool call is correct.\footnote{Note that HRE and MICE LR use a similar number of parameters, but in HRE they are devoted to closer analysis (binning) of the raw confidence dimension, rather than to additional BERTScore dimensions.}

\section{Results}
\label{sec:results}

\paragraph{Smooth Expected Calibration Error} 
Lower smECE is better.  
The first numeric column of Table~\ref{tab:results} shows that all of the confidence estimators are well-calibrated---their smECE values are small and not significantly different---except for the raw confidences, which have smECEs 3--10x higher than the others.   This is not surprising: HRE and {\kernelReg} are explicitly designed to calibrate the raw confidences, while logistic regression and random forest training are known to produce well-calibrated classifiers \cite{niculescu2005predicting}.  
\paragraph{\metricNameCaps} 
\Cref{fig:enr-plot-with-zeroshot} shows the \metricName curve for each confidence estimator and each model.
We find that raw confidence performs dangerously poorly at and above moderate risk levels. %
HRE and \kernelReg\ both degrade quickly toward
0
as risk increases.
The MICE models also degrade,
but more slowly:
matching performance of HRE and \kernelReg\ at the lower risk levels and outperforming at medium and higher risk levels.
Across all three LLMs, MICE RF performs best at nearly every risk level.  %

\Cref{tab:results} displays how well confidence estimators perform at three specific risk level settings (low, medium, and high) and across the full range of risk levels using AUC (see \cref{sec:metrics:reward}).
For all of these metrics, a higher score is better.
For each risk level, MICE RF always has the highest reward, outperforming HRE, \kernelReg, and MICE LR. 
Raw token confidence nearly always performs worst. 
For lower risk levels, most strategies perform comparably, with relatively high reward.
This is expected: executing an incorrect tool call ($\fp$) gives a low penalty relative to a correct tool ($\tp$), so aggressively biasing for execution is optimal, garnering a high reward. 
As risk levels increase, the penalty for executing an incorrect tool call grows and using raw confidences nearly always incurs a negative reward when the risk level is greater than 0.5 ($\fp < -1$).
Across the three risk levels, we find that the MICE models outperform both baselines for each of the three tool-calling LLM agents.

We run permutation tests for each metric in~\Cref{tab:results} for each MICE method as compared to HRE and \kernelReg.
In summary, MICE RF is always significant (p-value < 0.05) at the medium risk level, never significant at the low risk level, and significant at the high risk level for Llama3 and 3.1, but not 3.2. MICE LR outperforms the baselines, but is only significant for the medium risk level for Llama3.1.
For the summary statistic \AUC, both MICE models are nearly always significantly better than HRE and \kernelReg\ for all three Llamas.

\paragraph{Zero-Shot Generalization to New APIs} 

To test MICE's out-of-domain generalization, we simulate encountering new APIs by holding one out during training.
Since there are 50 APIs present in the STE dataset, we train 50 MICE RF and 50 MICE LR models.
Each model is trained on data from 49 APIs and evaluated solely on the held-out API. We combine the predictions from each of the models to get predictions across the entire test set.\footnote{This resembles 50-fold cross validation, where each fold is constructed solely with data from one API.
However, for comparability with other methods, we evaluate on the corresponding fold in the test set, not the training set.}
These confidence estimates are solely constructed by MICE models that have never seen that specific API before, so every tool is unseen. MICE does worse in this setting, but only degrades to the level of HRE and \kernelReg\ models trained on the full data; they are statistically indistinguishable from them.\looseness=-1 %

\section{Analysis}
\label{sec:analysis}

\paragraph{What is generated by decoding from intermediate layers?} Here, we look at what the LLM generates from intermediary layers. Using the logit lens, we find that models slowly evolve their predictions throughout the layers to get closer to the final output generation. 
\Cref{fig:gens-per-layer} shows sample generations.  
Qualitatively, the first two-thirds of the layers tend to generate seemingly random strings. After this point, the generations get increasingly closer to the final generation, but significant refinement still occurs in the final layer. %

The box-and-whisker plot in \Cref{fig:bertscores-per-layer} shows that BERTScore tends to increase with layer number.
\Cref{fig:bertscores-posneg} in \cref{sec:experimental-details} shows that at some layers, the distribution of BERTScores tends to be shifted \emph{slightly} higher on correct outputs, providing signal to the classifier.

\begin{figure}[t]
    \centering
    \includegraphics[width=\linewidth]{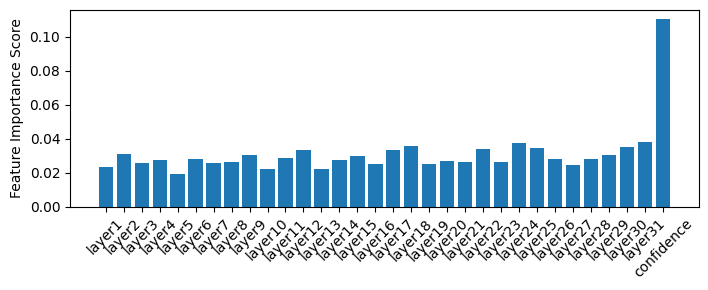}
    \caption{Feature importance for BERTScore features and confidence on the trained MICE RF model on the STE dataset for the Llama3 LLM.}
    \label{fig:rf-importance}
\end{figure}

\begin{figure}[t]
    \centering
    \includegraphics[width=\linewidth]{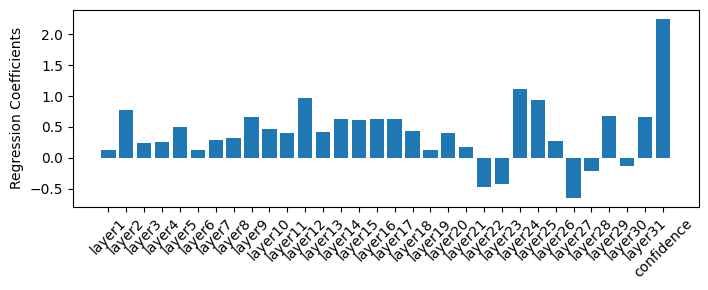}
    \caption{Coefficients for the trained MICE LR model on the STE dataset for the Llama3 LLM.}
    \label{fig:lr-importance}
\end{figure}

\paragraph{What is learned?}
To better understand how the MICE features are used, we examine our MICE models trained on the STE dataset with Llama3-8B-Instruct. 
For MICE RF we calculate Gini coefficients, and for MICE LR we analyze the feature weights, as suggested by reviewers.
\Cref{fig:rf-importance,fig:lr-importance} indicate that confidence is the most important feature in both MICE models: roughly 3 times as important as other features in MICE RF and 2 times as important as other features in MICE LR.\footnote{Perhaps calibrated confidence would have worked even better as a feature.} 
There is no obvious other pattern in the estimated weights, and it is possible that they are underdetermined.

\paragraph{Feature Ablations}
To better understand which features contribute most to confidence estimation, we performed feature ablations for both MICE models for the three LLMs in our study.
In addition to the original setting with all features, we tested four new settings:
confidence only;\footnote{For LR, this is exactly Platt scaling with L2 regularization.}
first half of the layers' BERTScores plus confidence;
second half of the layers' BERTScores plus confidence;
and all of the layers' BERTScores, but no confidence.
See Table~\ref{tab:results-ablation} in Appendix~\ref{sec:experimental-details} for details.

\textit{MICE RF:} Confidence alone performed extremely poorly.
The second half of the layers plus confidence performed better than the first half plus confidence, but using all layers without confidence performed worse than using half the layers with confidence.
This suggests that features from the second half of the model are more useful than the first half, and confidence is an important feature.

\textit{MICE LR:} Confidence alone performed comparably to other settings, indicating that confidence accounts for much of the performance; unlike RF, LR learned how to use this feature.
Additionally, for Llama3 and 3.1, using confidence alone outperformed using all the layers' BERTScore features.
Using the second half of the layers' BERTScore features outperformed using the first half of the layers' features, similar to MICE RF.

\begin{figure}[t]
    \centering
    \includegraphics[width=\linewidth]{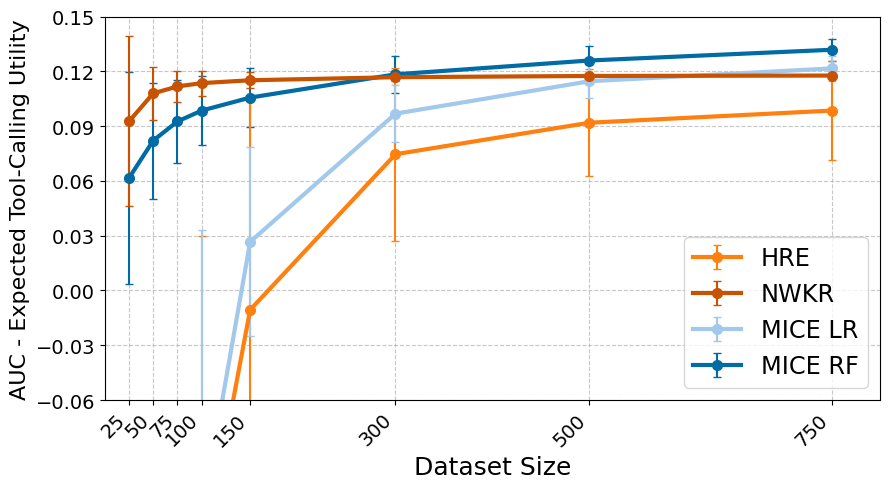}
    \caption{Sample complexity: AUC of MICE models and HRE baselines as the size of the training set varies on the Llama3-8B-Instruct model.
    Error bars are one standard deviation.
    }\label{fig:sample-complexity}
\end{figure}

\paragraph{How sample efficient is MICE?}
To measure sample efficiency, we vary the size of the training set to be 25, 50, 75, 100, 300, 500, and 750.
For each size, we randomly partition the 1,500 training examples into disjoint groups of that size (e.g., 15 groups of 100 examples, or 3 groups of 500 examples).
We then train on each group, measure \AUC, and compute the mean and variance across groups. %
We repeat this 100 times and average across trials, plotting results in~\Cref{fig:sample-complexity}.

For dataset sizes of 150 or below \kernelReg\ performs best, but it saturates at this level and does not improve further with more data.
MICE LR and HRE perform poorly with small datasets, but as size increases, they get closer to MICE RF and \kernelReg. %
For larger dataset sizes, MICE RF and MICE LR overtake \kernelReg.
In fact, with as few as 300 examples (20\% of the training data), MICE RF outperforms \kernelReg\ trained on the full dataset.

\section{Related Work}
\label{sec:related-work}

\paragraph{Model Internals}
\citet{tenney-etal-2019-bert} show that different layers of models encode different aspects of the classical NLP pipeline. Moreover, intermediate layer activations can be nudged via steering vectors to control output generations~\citep{subramani2019can, subramani-etal-2022-extracting, turner2023activation}. The activation spaces of models are relatively well-formed and there exist directions in these latent spaces that correlate with interpretable properties~\citep{subramani2020discovering, li2024inference}. 
These act as part of the basis for our hypothesis that the model internals could contain a trustworthiness signal, although we did not attempt to discover specific directions in these spaces. 

\paragraph{Intermediate Decoding}
We can view language models with multiple layers as doing iterative inference, where each successive layer refines the predictions of the previous layer.
With this lens, decoding from intermediate layers provides signal albeit noisy: the first half of the layers generate uninterpretable text, but after this predictions refine
toward a plausible answer~\citep{Belrose2023ElicitingLP, yom-din-etal-2024-jump, merullo-etal-2024-language}.
Other work has focused on inference efficiency by early exiting from transformers~\citep{teerapittayanon_branchynet_2017, geva-etal-2022-transformer, schuster2022confident, elhoushi2024layer}. 
Our work decodes from intermediary layers as a signal for better calibration. 

\paragraph{Calibration}

Prior work has measured the calibration of off-the-shelf models,
including neural networks~\cite{niculescu2005predicting,wang2024calibrationdeeplearningsurvey}, large language models~\citep{kadavath_language_2022,yin_large_2023}, and semantic parsers~\citep{stengel-eskin-van-durme-2023-calibrated,zhou-etal-2022-online}.

A line of machine learning work focuses on calibrating binary classifiers while conditioning only on their predicted confidence.
Platt scaling transforms a real-valued output (like that of an SVM classifier) into probabilities using logistic regression~\cite{Platt1999}, which is proven to be equivalent to beta calibration up to preprocessing~\cite{BOKEN2021101641}.
Isotonic regression~\cite{Ayer1955ANED} is a non-parametric approach that learns a best fit to data making only a monotonic non-decreasing assumption.
HREs are popular, and there has been work on adaptive binning strategies~\cite{nobel_histogram_regression_1996}.
We chose HRE and \kernelReg\ as strong baselines from this class of models.
MICE LR could be viewed as an extension to Platt scaling because MICE conditions on model internals in addition to the original confidence.

\paragraph{Applications of Well-Calibrated Confidences}
The \texttt{DidYouMean} system can rephrase a query and ask for confirmation when the model is unconfident~\cite{stengel-eskin-van-durme-2023-mean-fixed}.
Like us, they frame the competing concerns in terms of safety and utility, weighing wrongly predicted actions against the cost of asking clarifying questions.
While they tune a single confidence threshold, we transform confidences into calibrated probabilities so that a Bayes-optimal threshold can be dynamically derived for any risk/reward ratio.
LACIE~\cite{stengeleskin2024lacielistenerawarefinetuningconfidence} communicates its fine-tuned confidences to users.
APEL~\cite{zhong-etal-2023-non} reduces its uncertainty about a semantic parse by asking questions of a user, using \emph{raw} confidences to identify informative questions; calibrated confidences should work better, allowing it to finish with fewer questions.

\section{Conclusion}
\label{sec:conclusion}

In this work, we introduce \mice (MICE), which improve the trustworthiness and safety of language models as tool-calling agents.
We introduce a new metric, \metricName, that combines calibration and usefulness to better evaluate the safety and utility of tool calls.
We show that MICE matches or beats both regression baselines (HRE and NWKR) when measured by smooth ECE, and significantly improves \metricName, especially in medium and high-risk regimes.
Finally, we find that MICE %
is sample efficient and can generalize to unseen APIs in a zero-shot setting.

\section{Limitations}
\label{sec:limitations}

Like \logitLens, MICE assumes a transformer language model whose intermediate layers have the same shape as the final layer.  
More generally, MICE requires access to model internals, ruling out some of the most capable current LLMs, which are closed. %

In principle, MICE is a general-purpose confidence estimation recipe for transformer language models.  However, we evaluated MICE exclusively in one setting: a tool-calling task on one dataset. Other settings such as machine translation and question answering (see \cref{fn:othertasks}) have been left to future experiments.

As \cref{fn:altfeatures} hinted, there are many other
possible ways to compute MICE features.  We do not claim to have found the best variant even for the setting we studied.  While we settled on BERTScore for this paper, there are several other possible choices for how to \emph{encode}, \emph{align}, \emph{compare}, and \emph{aggregate} the tokens at each layer.  We remark that one possible encoding trick would be to learn a linear transform of each layer $i$ so that $\hiddenTokenAtLayer{i}{t-1}$ is maximally similar to $\hiddenTokenAtLayer{\ell}{t-1}$ or maximally predictive of $y_t$, as in the \emph{tuned lens} of \citep{Belrose2023ElicitingLP}.

There are also various ways to build a classifier that uses MICE features.  We also experimented with SVMs with different kernels (not reported).  Other options could also be tried.

\section{Impact Statement}
\label{sec:impact}

Better calibrated models can help people make safer decisions.
We hope to bring increased focus to risk/reward tradeoffs; we have intentionally framed the task and metric in a way that highlights the cost of false positives.
Decision theory and reward functions are not a substitute for careful design, however;
practitioners must exercise great care before hooking up an LLM to a tool with real effects in the world, including taking care to set appropriate rewards such as $\tp,\fp,\tn,\fn$.%

\bibliography{references,anthology_0,anthology_1}
\clearpage
\newpage

\appendix
\onecolumn

\section{Feature Ablation \& Analysis}
\label{sec:experimental-details}

\begin{table}[h]
\small
\begin{tabular}{@{}l l r r r r r@{}}
\toprule
 & & & \multicolumn{4}{c}{\textbf{Tool-Calling Utility} ($\uparrow$)}\\
\cmidrule(r){4-7}
\textbf{Base LLM} & \textbf{MICE Model}                    & \textbf{smECE} ($\downarrow$) & Low risk & Medium risk & High risk & AUC \\
\midrule
\multirow{10}{*}%
 & RF Confidence Only & 0.186 & 0.330 & -0.037 & -0.217 & 0.003 \\
 & RF First Half Layers + Confidence & \underline{0.035} & 0.353 & 0.065 & -0.007 & 0.127  \\
 & RF Second Half Layers + Confidence & 0.036 & {\bf 0.357} & \underline{0.097} & \underline{0.019} & \underline{0.143} \\
 & RF All Layers & 0.044 & \underline{0.356} & 0.081 & -0.005 & 0.129 \\
\rowcolor{LightCyan}\cellcolor{white}
Llama3-8B-I 
& RF & 0.040 & \underline{0.356} & {\bf 0.100} & {\bf 0.024} & {\bf 0.144}  \\
 & LR Confidence Only & 0.037 & \underline{0.356} & 0.035 & 0.000 & 0.120 \\
 & LR First Half Layers + Confidence & {\bf 0.034} & \underline{0.356} & 0.043 & 0.000 & 0.122 \\
 & LR Second Half Layers + Confidence & 0.041 & \underline{0.356} & 0.048 & 0.000 & 0.124 \\ 
 & LR All Layers & 0.055 & \underline{0.356} & 0.039 & 0.000 & 0.108 \\
\rowcolor{LightCyan}\cellcolor{white}
 & LR & 0.037 & \underline{0.356} & 0.055 & 0.000 & 0.125  \\
\midrule
\multirow{10}{*}%
 & RF Confidence Only & 0.183 & 0.182 & -0.097 & -0.061 & -0.012 \\
 & RF First Half Layers + Confidence & 0.039 & 0.242 & 0.057 & 0.003 & 0.086 \\
 & RF Second Half Layers + Confidence & {\bf 0.031} & \underline{0.247} & \underline{0.067} & 0.017 & \underline{0.098} \\
 & RF All Layers & {\bf 0.031} & 0.243 & 0.051 & \underline{0.020} & 0.095 \\
\rowcolor{LightCyan}\cellcolor{white}
Llama3.1-8B-I 
& RF & \underline{0.032} & {\bf 0.248} & {\bf 0.072}  & {\bf 0.023}  & {\bf 0.104}  \\
 & LR Confidence Only & 0.043 & 0.239 & 0.016 & 0.000 & 0.064 \\
 & LR First Half Layers + Confidence & 0.045 & 0.239 & 0.028 & 0.000 & 0.065 \\
 & LR Second Half Layers + Confidence & 0.050 & 0.239 & 0.049 & 0.000 & 0.070 \\
 & LR All Layers & 0.047 & 0.239 & 0.000 & 0.000 & 0.061 \\
\rowcolor{LightCyan}\cellcolor{white}
 & LR & 0.050 & 0.239 & 0.051  & \underline{0.000} & 0.071 \\
\midrule
\multirow{10}{*}%
 & RF Confidence Only & 0.158 & 0.196 & -0.065 & 0.005 & 0.010 \\
 & RF First Half Layers + Confidence & 0.034 & \underline{0.236} & 0.064 & \underline{0.011} & \underline{0.089} \\
 & RF Second Half Layers + Confidence & 0.041 & 0.235 & 0.059 & 0.009 & 0.084 \\
 & RF All Layers & 0.046 & \underline{0.236} & {\bf 0.075} & {\bf 0.013} & {\bf 0.095} \\
 \rowcolor{LightCyan}\cellcolor{white}
Llama3.2-3B-I 
& RF & 0.041 & {\bf 0.237} & \underline{0.071}  & {\bf 0.013} & {\bf 0.095} \\
 & LR Confidence Only & {\bf 0.022} & 0.233 & 0.000 & 0.000 & 0.057 \\.
 & LR First Half Layers + Confidence & 0.033 & 0.231 & 0.023 & 0.000 & 0.072 \\
 & LR Second Half Layers + Confidence & 0.026 & 0.233 & 0.012 & 0.000 & 0.064 \\
 & LR All Layers & 0.044 & 0.233 & 0.037 & 0.000 & 0.071 \\
\rowcolor{LightCyan}\cellcolor{white}
 & LR & \underline{0.025} & 0.232 & 0.016 & 0.000 & 0.073  \\
\bottomrule
\end{tabular}
\caption{%
Results of the feature ablation on the STE test set~\cite{wang-etal-2024-llms-imaginarium}.
Lower smECE is better, while higher tool-calling utility is better.
\textbf{Bold} indicates the best result in each category and \underline{underline} indicates the second best result in each category.
}
\label{tab:results-ablation}
\end{table}

\begin{figure}[h!]
\includegraphics[width=\linewidth]{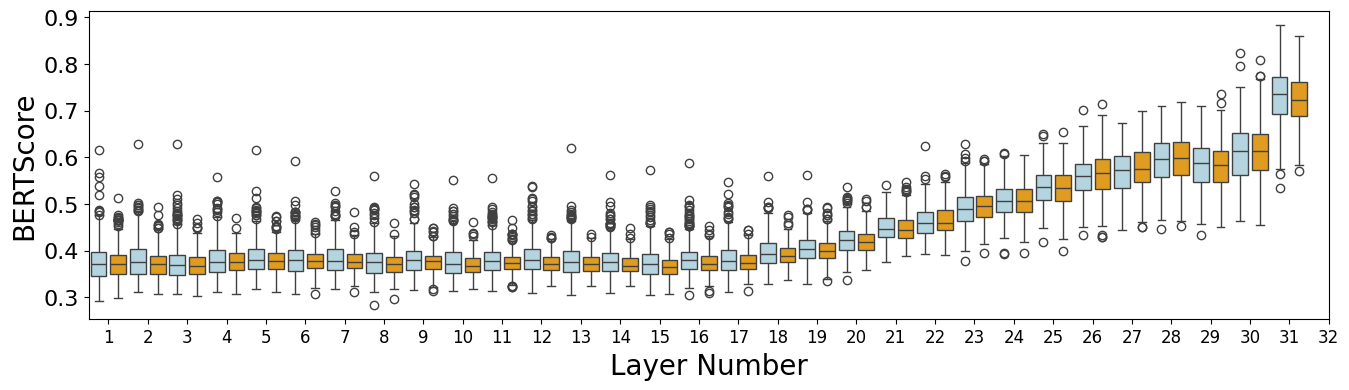}
\caption{A version of \cref{fig:bertscores-per-layer} that shows how positive examples (blue) and negative examples (orange) may have slightly different BERTScore distributions at each layer (see \cref{sec:analysis}), which is apparently enough to inform the confidence estimator.  In both figures, we use the standard boxplot function in the Seaborn package with default hyperparameters~\citep{Waskom2021}.}\label{fig:bertscores-posneg}
\end{figure}

\end{document}